\documentclass{bmvc2k}
\usepackage{multirow}
\usepackage{wasysym}
\usepackage{tabularx}
\usepackage{amsmath}
\usepackage{wrapfig}
\usepackage{hhline}
\usepackage{tablefootnote}
\usepackage{marvosym}

\title{DA-CIL: Towards Domain Adaptive Class-Incremental 3D Object Detection}

\addauthor{Ziyuan Zhao}{zhao_ziyuan@i2r.a-star.edu.sg}{1,2,\Letter}
\addauthor{Mingxi Xu}{XU0002XI@e.ntu.edu.sg}{1,3}
\addauthor{Peisheng Qian}{qian_peisehng@i2r.a-star.edu.sg}{1}
\addauthor{Ramanpreet Singh Pahwa}{ ramanpreet_pahwa@i2r.a-star.edu.sg}{1,2}
\addauthor{Richard Chang}{richard_chang@i2r.a-star.edu.sg}{1}

\addinstitution{
 Institute for Infocomm Research (I$^2$R),
 Agency for Science, Technology and Research (A*STAR), Singapore
}
\addinstitution{
 Artificial Intelligence, Analytics And Informatics (AI$^3$),
 Agency for Science, Technology and Research (A*STAR), Singapore
}
\addinstitution{
 Nanyang Technological University, Singapore
}

\runninghead{ZHAO ET AL}{DA-CIL}

\begin{document}

\maketitle

\begin{abstract}
Deep learning has achieved notable success in 3D object detection with the advent of large-scale point cloud datasets. However, severe performance degradation in the past trained classes,~\emph{i.e.}, catastrophic forgetting, still remains a critical issue for real-world deployment when the number of classes is unknown or may vary.
Moreover, existing 3D class-incremental detection methods are developed for the single-domain scenario, which fail when encountering domain shift caused by different datasets, varying environments, etc. 
In this paper, we identify the unexplored yet valuable scenario,~\emph{i.e.}, class-incremental learning under domain shift, and propose a novel 3D domain adaptive class-incremental object detection framework, DA-CIL, in which we design a novel dual-domain copy-paste augmentation method to construct multiple augmented domains for diversifying training distributions, thereby facilitating gradual domain adaptation. 
Then, multi-level consistency is explored to facilitate dual-teacher knowledge distillation from different domains for domain adaptive class-incremental learning.
Extensive experiments on various datasets demonstrate the effectiveness of the proposed method over baselines in the domain adaptive class-incremental learning scenario.
\end{abstract}


\section{Introduction}
\label{sec:intro}
With the widespread use of 3D point clouds, deep learning-based 3D object detection has received considerable attention.
Many efforts~\cite{qi2017pointnet, qi2017pointnet++, qi2019deep, pham20203d, pahwa2021automated, nwe2020improving} have been devoted to the field, achieving remarkable success in object recognition and localization from 3D point clouds or voxelized data.
Nevertheless, the~\emph{catastrophic forgetting} problem seriously limits the deployment of existing detection models in dynamic real-world environments, where novel classes are encountered over time.
In the class-incremental scenario, the performance of existing models on old classes tends to decrease substantially when trained on novel classes. 
In recent years, class-incremental learning has been extensively investigated for various 2D vision tasks from different perspectives, including regularization-based methods~\cite{hou2019learning}, replay-based methods~\cite{rebuffi2017icarl, lopez2017gradient} and parameter-isolation methods~\cite{mallya2018packnet, serra2018overcoming}. However, class-incremental learning (CIL), particularly class-incremental 3D object detection remains underexplored~\cite{dong2021i3dol,zhao2022static}. 

\begin{figure*}[t]
\centering
\includegraphics[width=0.9\textwidth]{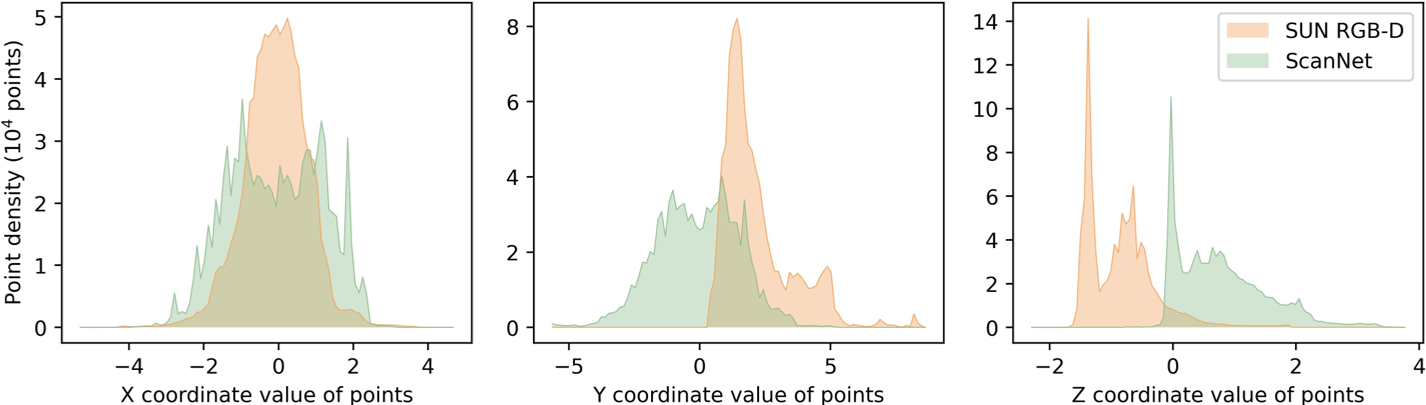}
\caption{Distribution of object dimensions in SUN RGB-D~\cite{song2015sun} and ScanNet~\cite{dai2017scannet}.}
\label{fig:dataset_distribution}
\end{figure*}

On the other hand, deep learning models are also required to rapidly adapt to varying environmental conditions, such as locations, surroundings, and weather.
Existing class-incremental 3D object detection methods assume that the data is from the same domain with the same distribution. 
In the presence of domain shift, deep learning models trained on one domain (\emph{i.e.,} source domain) always suffer from tremendous performance degradation when evaluated on another domain (\emph{i.e.,} target domain). Moreover, multiple datasets are encouraged to be used to alleviate data scarcity in a cross-domain scenario, inducing the notorious domain gap across datasets by various factors, such as geometric mismatch (Fig.~\ref{fig:dataset_distribution}).
A series of unsupervised domain adaptation~(UDA) approaches have emerged to address the domain shift in various computer vision tasks~\cite{tsai2018learning,hoffman2018cycada,chen2021gradual,dai2021idm}.
However, these methods are not tailored to meet the needs of class-incremental object detection from 3D point clouds. 

\begin{figure*}[t]
\centering
\includegraphics[width=0.9\textwidth]{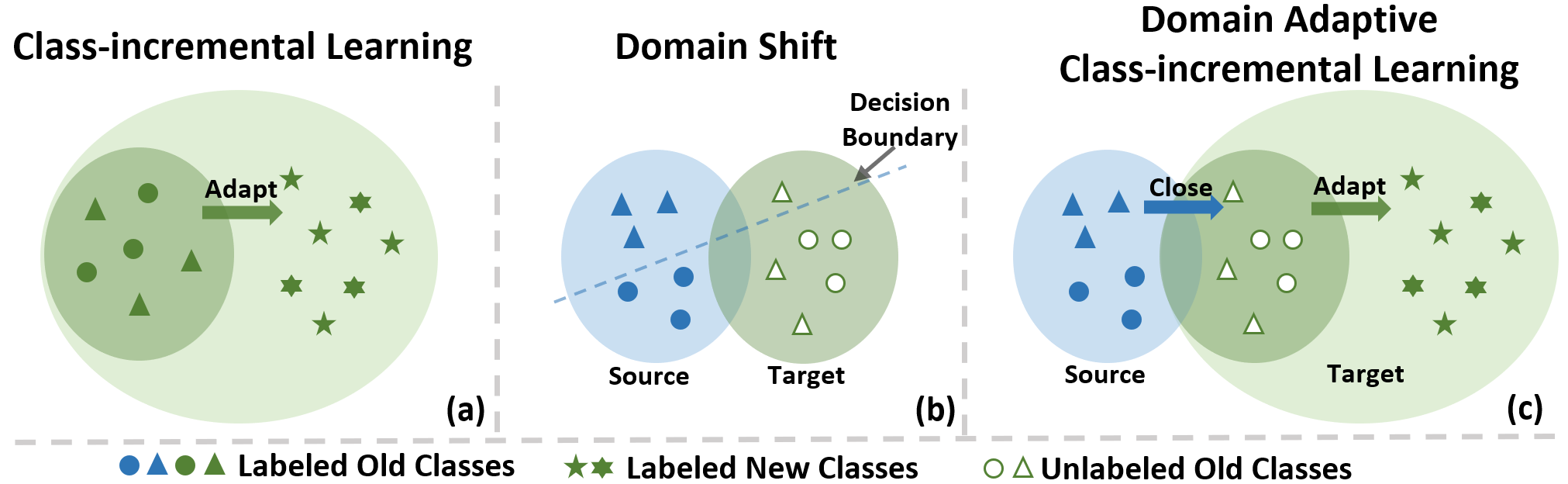}
\caption{The difference between CIL and DA-CIL. (a) CIL is designed for adapting new classes without forgetting old classes in the same domain. (b) Domain shift always exists between source and target domains. (c) DA-CIL is designed to close the domain shift and adapt new classes across different domains.}
\label{fig:scenario}
\end{figure*}

In summary, existing CIL and UDA methods are designed for addressing different challenges separately, ignoring that the \emph{catastrophic forgetting} and \emph{domain shift} problems concurrently exist in 3D object detection on point clouds of real-world environments. This motivates us to investigate a practical yet challenging class-incremental scenario,~\emph{i.e.,} class-incremental learning under domain shift, and introduce a new CIL paradigm called domain adaptive class-incremental learning~(DA-CIL). 
To the best of our knowledge, we are~\textit{the first} to study domain adaptation in class-incremental learning for 3D object detection. 
As shown in Fig.~\ref{fig:scenario}, we aim to leverage labeled old classes on the source domain and labeled new classes on the target domain to close the domain shift and adapt to new classes without forgetting old classes on the target domain in DA-CIL. 
We propose a novel DA-CIL framework for 3D object detection from point clouds. 
First, we propose a dual-domain copy-paste (DuDo-CP) augmentation method to generate \textbf{cross-domain} point clouds and \textbf{in-domain} point clouds for ground-truth augmentation and gradual domain adaptation to narrow the distribution shift across domains. 
Second, we employ a dual-teacher network to facilitate knowledge transfer from different domains. The cross-domain teacher trained with cross-domain point clouds is used to generate pseudo labels of old classes in the target domain and distill cross-domain knowledge of old classes to the student model. The in-domain teacher based on self-ensembling~\cite{tarvainen2017mean} is used to transfer in-domain knowledge of new classes to the student model. 
Third, we construct multi-level consistency (MLC) regularization from different aspects to better facilitate the dual-domain teacher-student training process. Our approach is broadly applicable to various applications,~\emph{e.g.}, autonomous driving, robotics and interior design. For example, when autonomous driving models developed in one domain are applied to a new domain with different distributions (e.g., new roads and different light conditions) and
novel classes (e.g., local vehicles and animals), the proposed
framework can help effectively tackle both~\emph{domain shift} and~\emph{catastrophic forgetting}.

Our main contributions can be summarized as follows: 
  
\begin{itemize}
\item  We identify a new CIL scenario where domain shift occurs when adapting new classes across domains, and formulate a new CIL paradigm to enable domain adaptive class-incremental learning for 3D object detection.

\item  Extensive experiments and analysis demonstrate that our approach can achieve superior performance over existing methods under different incremental learning scenarios.
\end{itemize}

\section{Related Work}
\noindent\textbf{Point cloud object detection.} 
3D object detection from point clouds has drawn great attention in recent years. PointNet~\cite{qi2017pointnet} is a pioneering work to detect 3D objects using only point cloud data, which extracts global information and point-wise features for 3D object detection.
PointNet++~\cite{qi2017pointnet++} advances PointNet by extracting local information in neighborhoods of individual points at multiple scales. 
More recently, Qi~\emph{et al.} propose a voting-based framework, VoteNet~\cite{qi2019deep}, in which high-quality object bounding boxes are generated by directly voting for object centers, achieving promising performance on various indoor benchmarks such as ScanNet~\cite{dai2017scannet} with geometric information alone. Therefore, we use the modified VoteNet~\cite{zhao2022static} as our detection backbone.

\noindent\textbf{Class-incremental learning.} 
Class-incremental learning aims to alleviate the \emph{catastrophic forgetting} effect and enable deep learning models to continuously learn new classes while preserving the learned knowledge from old classes.
Most of the existing class-incremental learning methods were developed for 2D planar datasets, which can be divided into three categories~\cite{de2021continual}: 
1) \emph{Regularization-based methods} use knowledge distillation~\cite{hinton2015distilling} to enforce consistency between either the data (\emph{e.g.,} rebalancing~\cite{hou2019learning}) or parameters (\emph{e.g.,} EWC~\cite{kirkpatrick2017overcoming}).
2) \emph{Replay-based methods} such as iCaRL~\cite{rebuffi2017icarl} and GEM~\cite{lopez2017gradient} store exemplars from old classes in a replay buffer or reproduce samples of previous tasks with a trained generator.
3) \emph{Parameter isolation methods} can add new branches for new classes while freezing model parameters learned from old classes, for instance, PackNet\cite{mallya2018packnet} and HAT~\cite{serra2018overcoming}.
In comparison, only a few methods explore 3D class-incremental scenarios, especially for class-incremental 3D object detection~\cite{dong2021i3dol, zhao2022static}. For example, SDCoT~\cite{zhao2022static} is a co-teaching method that alleviates catastrophic forgetting of old classes via a static teacher, and consistently learns the underlying knowledge from new data via a dynamic teacher. However, these methods focus on the single-domain scenario and cannot handle the domain shift issue.


\noindent\textbf{Unsupervised domain adaptation.} 
Domain shift has been a long-standing problem for computer vision tasks, especially in the absence of target labels~\cite{wang2018deep}. 
Considerable efforts have been devoted to studying unsupervised domain adaptation (UDA) for minimizing the cross-domain discrepancy from different perspectives, including adversarial learning~\cite{tsai2018learning,hoffman2018cycada,jie2022adan}, self-ensemble learning~\cite{zhao2021mt,luo2021unsupervised,zhao2022meta}, and self-training~\cite{zou2019confidence,yang2021st3d}. Nevertheless, limited studies have been proposed for UDA in 3D object detection. Wang~\emph{et al.}~\cite{wang2020train} proposed to align the object size across domains to close the size-level domain gap. Luo~\emph{et al.}\cite{luo2021unsupervised} proposed a Multi-Level Consistency
Network (MLC-Net) based on the mean teacher paradigm~\cite{tarvainen2017mean} for addressing the geometric-level domain gap. These UDA methods require samples from both source and target domains for domain alignment, thereby having limited use on data-scarce scenarios. Besides, the same classes are required for both domains in conventional UDA, limiting the practical use for class-incremental scenarios. Recent works~\cite{dai2021idm, ramamonjison2021simrod} suggest that augmentation techniques~\cite{yun2019cutmix, nekrasov2021mix3d,ghiasi2021simple} can not only relieve data scarcity but also remedy data-level gaps across domains. Inspired by the success of Copy-Paste augmentation~\cite{ghiasi2021simple}, we intend to advance it on cross-domain scenarios to diversify training data distributions for domain adaptive class-incremental learning.






\begin{figure*}[t]
\centering
\includegraphics[width=0.95\textwidth]{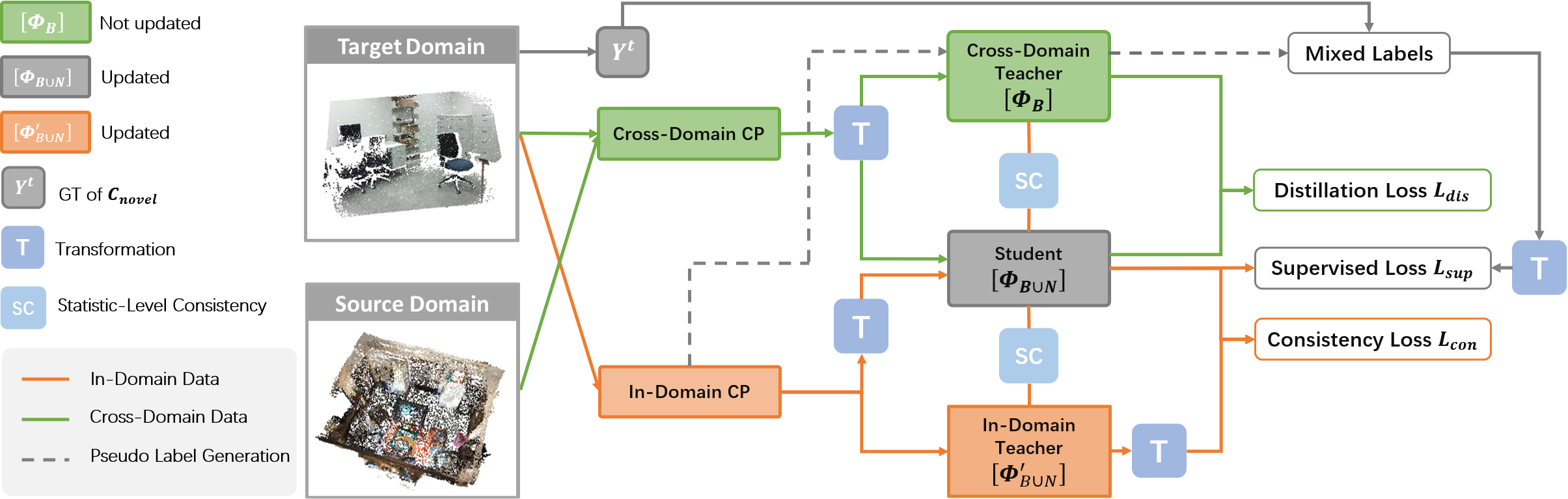}
\caption{Overview of the proposed DA-CIL framework: First, cross-domain intermediate point clouds and in-domain augmented point clouds are generated by dual-domain copy-paste augmentation. Then, a dual-teacher network is constructed to distill knowledge from these domains for domain adaptive class-incremental learning.}
\label{fig:architecture_overview}
\end{figure*}

\section{Methodology}

Let $C=C_{b} \cup C_{n}$ denote the whole set of classes, where $C_{b}$ is the set of base (old) classes and $C_{n}$ is the disjoint set of novel (new) classes. In the conventional class-incremental object detection,  $C_{b}$ and  $C_{n}$ have great amounts of training samples from the same domain with annotated object classes and bounding boxes. Differently, in our DA-CIL setting, $C_{b}$ and  $C_{n}$ are from different domains with distribution shift. We define the set of base classes from the source domain as $C^{s}_{b}$, and the set of novel classes from the target domain as $C^{t}_{n}$. Given a well-trained base model $\Phi_B$ on $C^{s}_{b}$, we aim to learn a domain adaptive class-incremental model $\Phi_{B\bigcup N}$ using $C^t_n$, which can detect both base classes $C^{t}_{b}$ and novel classes $C^{t}_{n}$ in the target domain. In this regard, we propose a unified DA-CIL framework to address both~\emph{catastrophic forgetting} and~\emph{domain shift} problems simultaneously.

\subsection{DA-CIL Architecture\label{architecture}}

Fig.~\ref{fig:architecture_overview} depicts an overview of the proposed DA-CIL architecture. In DA-CIL, two teacher-student networks are constructed for dual-domain knowledge distillation to achieve class-incremental learning and domain adaptation. More specifically, we first generate cross-domain intermediate point clouds and in-domain augmented point clouds using the proposed dual-domain copy-paste (DuDo-CP) augmentation (see Sec.~\ref{sec:copy-paste_augmentation}) to diversify training distributions, thereby improving the domain adaptive class-incremental performance. Then, we construct a dual-teacher network, which consists of one student model $\Phi_{B\bigcup N}$, one cross-domain teacher $\Phi_B$, and one in-domain teacher ${\Phi'}_{B\bigcup N}$. The well-trained cross-domain teacher on intermediate point clouds is applied to regularize the student model for closing domain shift and generate pseudo labels for mixed label creation. At the same time, the transformed and original in-domain augmented point clouds are passed to the student model and in-domain teacher model, respectively, for multi-level consistency learning  (see Sec.~\ref{sec:multi-level}). It is noted that the in-domain teacher is updated with the exponential moving average (EMA) weights of the student model,~\emph{i.e.,} ${\Phi'}_{B\bigcup N}^t = \alpha{\Phi'}_{B\bigcup N}^{t-1}+ (1-\alpha) \Phi_{B\bigcup N}^t$  where $t$ is the training iteration and $\alpha$ is the EMA decay rate. Finally, the student model is trained under mixed label supervision with dual-teacher knowledge transfer to achieve domain adaptive class-incremental learning (see Sec.~\ref{sec:training strategies}).

\subsection{Dual-Domain Copy-Paste}
\label{sec:copy-paste_augmentation}

\begin{figure*}[t]
    \centering\includegraphics[width=0.8\textwidth]{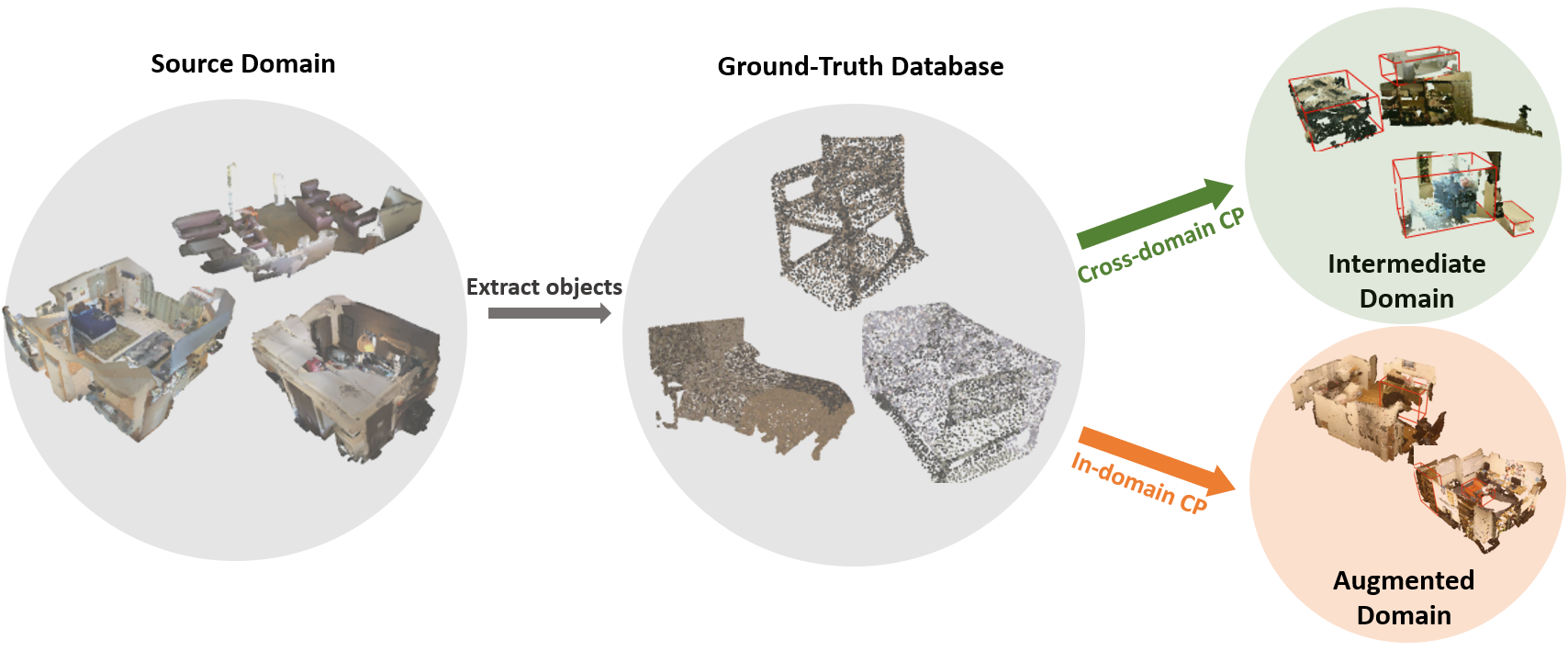}
\caption{An illustration of Dual-Domain Copy-Paste, which first extracts objects from the source domain to populate the ground-truth database. Then objects are copy-pasted into point clouds from the source domain (In-domain CP) and the target domain (Cross-domain CP) to generate the augmented domain and the intermediate domain, respectively. }

\label{fig:copy_paste}
\end{figure*}

To relieve data scarcity and reduce the domain gap at the data level, we extensively leverage copy-paste (CP) augmentation techniques for creating cross-domain and in-domain point clouds. As shown in Fig.~\ref{fig:copy_paste}, we generate a ground-truth database containing objects in $C^s_b$, and randomly select objects in the ground-truth database to paste in point clouds from source and target domains, forming in-domain source point clouds $X_{in(source)}$ and cross-domain point clouds $X_{cross}$, respectively. Similarly, $X_{in(target)}$ can be constructed by pasting $C^t_n$ objects into point clouds in the target domain. 

\noindent\textbf{Cross-domain CP.} To mitigate the serious domain shift between the source domain $X_s$ and the target domain $X_t$ caused by point cloud geometric mismatch (see Fig.~\ref{fig:dataset_distribution}), we construct the intermediate domain $X_{cross}$. More specifically, we explicitly model $X_{cross}$ that includes source objects in the target context. In $X_{cross}$, the distribution of context around source objects becomes consistent with $X_t$. Meanwhile, local geometries of objects remain unchanged. In $X_{cross}$, data distribution has shifted from $X_s$ to the combination of source objects in target surroundings, which is between $X_t$ and $X_s$ as shown in Fig.~\ref{fig:scenario}. Therefore, $X_{cross}$ serves as a bridge between $X_s$ and $X_t$ to facilitate the gradual transfer of knowledge~\cite{dai2021idm}. Let $D^s_i\in X_s, (i=1,2,3,...)$ denotes objects in the source domain $X_s$. We copy-paste $D^s_i$ into $X_t$ and construct an intermediate domain $X_{cross}$, formulated as:
\begin{equation}
\label{equ:intermediate_domain}
    X_{cross} = X_t + \sum_{n=1}^{N} T(D^s_n),
\end{equation}
where N refers to the number of objects, and T refers to object transformations, including random size and orientation augmentations. During training, $X_{cross}$ is first created to fine-tune the base model pre-trained on $X_s$. Subsequently, $X_{cross}$ is fed to the cross-domain teacher and the student in Fig.~\ref{fig:architecture_overview} for distillation of knowledge in old classes.

\noindent\textbf{In-domain CP.} To tackle data scarcity in $X_s$ and $X_t$, we perform in-domain copy-paste, which diversifies object appearances and their surrounding environment. Specifically, we create an augmented source domain $X_{in(source)}$ and an augmented target domain $X_{in(target})$, calculated as:
\begin{equation}
\label{equ:augmented_source_domain}
    X_{in(source)} = X_s + \sum_{n=1}^{N} T(D^s_n);~X_{in(target)} = X_t + \sum_{n=1}^{N} T(D^t_n),
\end{equation}
    
where $D^t_i\in X_t, (i=1,2,3,...)$ denotes objects in the target domain. The approach complements the cross-domain copy-paste and enhances the model performance in individual domains. During training, $X_{in(source)}$, and $X_{in(target)}$ are constructed for base model pre-training and dual-teacher knowledge distillation, respectively. 



\subsection{Multi-Level Consistency}
\label{sec:multi-level}
To facilitate the dual-teacher knowledge transfer, we propose multi-level consistency regularization from two aspects. First, we propose statistics-level consistency to tackle the mismatch in batch normalization statistics. Second, bounding box-level consistency (\emph{i.e.,} center-level, class-level and size-level consistency~\cite{zhao2020sess}) is applied for intra-domain knowledge distillation. 

\noindent\textbf{Statistics-Level Consistency.} 
In our architecture, the inputs of the student and teacher models are different, resulting in batch statistics mismatch and an unstable training process. The mismatched batch statistics, such as parameters in batch normalization (BN) layers (mean $\mu$ and variance $\sigma$) could degrade the model performance~\cite{cai2021exponential}. In this regard, we propose to share the BN parameters of the in-domain teacher with the cross-domain teacher and the student network for closing the distribution gap caused by statistics mismatch and stabilizing the training process.

\noindent\textbf{Bounding Box-Level Consistency.}
Let $B_s=\left\{b_s^1,b_s^2,...,b_s^N\right\}$ and $B_{in}=\left\{b_d^1,b_d^2,...,b_d^M\right\}$ be the proposals generated by the student and the in-domain teacher, respectively, where $N$ and $M$ refer to the respective number of proposals from student and in-domain teacher. Our bounding box-level consistency includes $3$ components: 1) \textbf{center-level consistency loss} calculates the distance between centers of bounding boxes proposed by the student and the in-domain teacher, shown in Eq.~\ref{equ:center_loss}, 2) \textbf{ class-level consistency loss} computes the Kullback-Leibler (KL) Divergence between class proposals from the student and the in-domain teacher, as shown in Eq.~\ref{equ:class_loss}, and 3) \textbf{size-level consistency loss} is the mean squared error (MSE) of widths, lengths, and heights of paired bounding boxes calculated in Eq.~\ref{equ:size_loss},  
\begin{equation}
\label{equ:center_loss}
    L_{center} = dist(Pair_N)+dist(Pair_M),
\end{equation}

\begin{equation}
\label{equ:class_loss}
    L_{class} = \frac{1}{M}\sum_{i=1}^M D_{KL}(p_s^i,p_d^j),\quad (b_s^i,b_d^j)\subset{Pair_M},
\end{equation}

\begin{equation}
\label{equ:size_loss}
    L_{size} = \frac{1}{M}\sum_{i=1}^M MSE(s_s^i,s_d^j),
\end{equation}
where $dist$ refers to the square of the Euclidean distance between the centers. $Pair_M$ is a paired bounding box set. Each pair contains an in-domain teacher proposal $b_d^i$ and the closest student proposal $b_s^j$ to $b_d^i$ based on the minimum Euclidean distance between their centers and same to $Pair_N$~\cite{zhao2020sess}.
$p_s^i$ is the probability of $b_s^i$ belonging to each class. $s_s^i$ contains the length, width and height of $b_s^i$, and the same setting applies for $b_d^i$.

\subsection{Dual-Teacher Training Strategies}
\label{sec:training strategies}

We implement the modified VoteNet~\cite{zhao2022static} as our detection backbone for both teacher and student models, in which the indices of sampled points and votes from the student model are stored and reused in the cross-domain teacher for class-incremental learning. The student model learns the underlying knowledge in base classes from a cross-domain teacher (\emph{i.e.,} a frozen model pre-trained on the intermediate domain $X_{cross}$ and augmented source domain $X_{in(target)}$) via a distillation loss $L_{dis}$, which is the square of Euclidean distance between the classification logits of different bounding boxes. 
Meanwhile, the EMA in-domain teacher helps student model capture structure and semantic invariant information in objects with a consistency loss $L_{con}$, shown as:

\begin{equation}
\label{equ:L_conf}
    L_{con} = L_{center} + \lambda_{class}L_{class} + \lambda_{size}L_{size},
\end{equation}
where $\lambda_{class}$ and $\lambda_{size}$ are both set to $1$, empirically.

The 3D bounding boxes generated by the cross-domain teacher for the base classes in the target domain $X_{in(target)}$ are utilized as pseudo labels which can be combined with labels of the novel classes to form the mixed labels for supervised learning.
More specifically, the mixed labels are transformed by the same augmentation step that is applied on $X_{in(target)}$ to compute a supervised loss $L_{sup}$, following multi-task loss in VoteNet~\cite{qi2019deep}. The final loss $L$ used for updating the student model is calculated as:

\begin{equation}
\label{equ:L_all}
    L = \lambda_{sup}L_{sup} + \lambda_{dis}L_{dis} + \lambda_{con}L_{con}, 
\end{equation}
where $\lambda_{sup}$, $\lambda_{dis}$, and $\lambda_{con}$ are set to $10$, $1$, and $10$, empirically.

\section{Experiments}

\subsection{Datasets and Evaluation Metrics}

We evaluate our method on 3D object detection datasets, ScanNet~\cite{dai2017scannet} and SUN RGB-D~\cite{song2015sun}. ScanNet is an RGB-D video dataset containing $1,513$ scans with object bounding boxes derived from point-level segmentation labels. SUN RGB-D is captured by $4$ different sensors and contains $10,335$ RGB-D images with $64,595$ 3D bounding boxes including orientations. To simulate the domain adaptive class-incremental scenario, ScanNet was employed as the source domain and $5$ categories (bathtub, bed, bookshelf, chair, desk) were selected as base classes $C^b_s$. SUN RGB-D was employed as the target domain, in which $C^b_t$ refers to the same $5$ base classes. 5 additional categories (dresser, nightstand, sofa, table, toilet) in SUN RGB-D were selected as novel classes in target domain,~\emph{i.e.}, $C^n_t$. For evaluation, we adopted the common metric~\cite{qi2019deep}, mAP@IoU=0.25,~\emph{i.e.}, mean average precision of predicted bounding boxes that overlap ground-truth with an amount of $0.25$ or higher. Higher values indicate better performance.

\subsection{Implementation Details}
\label{sec:implementation_details}

For DuDo-CP, we randomly pasted $1$ or $2$ $C^b_s$ object(s) to each scene in target domain $X_t$ to form cross-domain point clouds $X_{cross}$. The $C^b_s$ objects were randomly positioned on the left and right half along the x-axis of the point cloud to avoid collision. To address data scarcity, random scaling between $(0.9, 1.1)$ and random rotation of $\pm10$ degrees were applied to the objects. 
We first conducted base training on $C^b_s$ for 150 epochs and fine-tuned using $X_{in(source)}$ and $X_{cross}$ sequentially for $5$ epochs each. 
We used the Adam optimizer with a batch size of $8$ and an initial learning rate of $0.001$, decayed at epoch $80$ and $120$ by $0.1$. Then, we trained the dual-teacher network for $100$ epochs. The teacher and student networks were initialized using the base-train model weights. Following~\cite{tarvainen2017mean}, the EMA decay rate was set to $0.999$. In each batch, $6$ point clouds were selected from $X_{in(target)}$ and $2$ point clouds were selected from $X_{cross}$ for training. The remaining settings were the same as base training.

\subsection{Baselines}
We compared our results with the following baselines. VoteNet~\cite{qi2019deep} was used as the base model in all experiments. Following~\cite{zhao2022static}, we modified the VoteNet for aligning proposals from teacher and student models to facilitate class-incremental learning without leveraging color information. The model size is 10.9MB with 0.94M parameters. The inference time for a single 3D point cloud is around 0.2370s. It was trained on $C^b_s$ and $C^b_t$ to measure the domain gap in base classes. We then conducted joint training to evaluate the domain gap in all classes. For class-incremental learning, we evaluated $2$ naive transfer learning baselines~\textbf{Freeze-and-add}, and~\textbf{Fine-tune}. To be specific, \textbf{Freeze-and-add} adds a new classifier for novel classes on the pre-trained base model with frozen weights.~\textbf{Fine-tune} is similar to freeze-and-add except that all model parameters excluding the old classifier are not frozen. 
We implemented one recent popular self-ensembling method,~\textbf{MT}~\cite{tarvainen2017mean} for unsupervised domain adaptation, and pseudo labels of base classes generated by the base model were introduced for adapting MT to class-incremental scenarios.
We also used~\textbf{SDCoT}~\cite{zhao2022static}, the co-teaching architecture as a strong baseline for class-incremental learning.

\begin{table}[t]

\caption{3D object detection performance mAP@0.25 on the SUN RGB-D validation set.}
\centering
\setlength\tabcolsep{8pt}
\scalebox{0.8}{
\begin{tabular}{ |c|c|c|c|c|c|c|c| } 
 \hline
 & Method & $C_s^b$ & $C_t^b$ & $C_t^n$ & Base & Novel & All \\
 \hline
 \multicolumn{1}{|c|}{\multirow{2}{*}{Base only}}&Base train& \checked &$\times$&$\times$& 26.60 & - & -\\ \cline{2-8}
 &Base train&$\times$& \checked &$\times$& 57.63 & - & -\\ 
 \hhline{========}
 \multicolumn{1}{|c|}{\multirow{4}{*}{Class-Incremental Learning (CIL)}}&Freeze-and-add&$\times$& \checked & \checked & 54.24 & 10.61 & 32.42\\ \cline{2-8}
 &Fine-tune& $\times$ & \checked & \checked & 3.48 & 54.10 & 28.79\\ \cline{2-8}
 &MT~\cite{tarvainen2017mean}&$\times$& \checked & \checked & 49.63 & \textbf{61.21} & 55.42\\ \cline{2-8}
 &SDCoT~\cite{zhao2022static}&$\times$& \checked & \checked & 52.04 & 59.48 & 55.76\\ \cline{2-8}
 &Ours&$\times$& \checked & \checked & \textbf{53.81} & 59.50 & \textbf{56.66}\\
  \hhline{========}
  \multicolumn{1}{|c|}{\multirow{4}{*}{CIL under Domain Shift}}&Freeze-and-add& \checked &$\times$& \checked & 23.31 & 2.04 & 12.68\\ \cline{2-8}
 &Fine-tune& \checked &$\times$& \checked & 1.73 & 58.24 & 29.99\\ \cline{2-8}
 &MT~\cite{tarvainen2017mean}& \checked &$\times$& \checked & 33.84 & \textbf{59.03} & 46.43\\ \cline{2-8}
 &SDCoT~\cite{zhao2022static}& \checked &$\times$& \checked & 29.41 & 57.74 & 43.57\\ \cline{2-8}
 &Ours& \checked &$\times$& \checked & \textbf{36.41} & 58.80 & \textbf{47.60}\\
  \hhline{========}
  \multicolumn{1}{|c|}{\multirow{2}{*}{Base \& Novel}}&Joint train& \checked &$\times$& \checked & 6.80 & 52.52 & 29.66\\ \cline{2-8}
  &Joint train&$\times$& \checked & \checked & 58.11 & 59.27 & 58.69\\\hline

\end{tabular}
}
\label{tab:main_results}
\end{table}

\subsection{Experimental Results}
As shown in Table~\ref{tab:main_results}, the base train result with $C^b_s$ is significantly worse than that with $C^b_t$, which confirms the severity of the domain gap. SDCoT cannot well handle the domain shift problem, while MT improves the results, which demonstrate that self-ensemble learning can help domain adaptation in the class-incremental scenarios. Next, our method has a slight edge over SDCoT in the class-incremental learning (CIL) scenario, which proves the effectiveness of our approach in a single domain. Furthermore, our method outperforms SDCoT by a clear margin in CIL under domain shift. 
There is a $4\%$ performance improvement in all classes ($7\%$ on base classes), indicating the effectiveness of our DA-CIL framework. In addition, for the CIL scenario, mAP@0.5 is 30.36\% and 32.16\% for SDCoT and our method, respectively. For CIL under domain shift scenario, mAP@0.5 is 21.19\% and 22.39\% for SDCoT and our method, respectively. Besides, it is well noted that our results ($36.41\%$) have greatly improved from the lower bound ($26.60\%$) on base classes. Finally, the domain gap is confirmed between two results of joint training on $C_s^b~\bigcup C_t^b$, and $C_s^b~\bigcup C_t^n$ respectively. 
We further investigate model performance in each class under the domain adaptive class-incremental setting, as presented in Table~\ref{tab:per_class}. Our method achieves better results in most classes, especially for large objects such as bathtubs and beds. They have relatively fixed contexts, which can benefit more from the augmented object-context combinations. On the other hand, objects like chairs receive slight performance degradation because they are small and can be erroneously copy-pasted with other objects. Visualizations of object detection results on SUN RGB-D validation data are provided in the supplementary material.

\begin{table}[t]
\label{tab:incremental_learning_results}
\caption{Per-class object detection performance AP@0.25 on SUN RGB-D validation set, under the domain adaptive class incremental setting.}
\centering
\setlength\tabcolsep{3pt}
\scalebox{0.75}{
\begin{tabular}{ |c|c|c|c|c|c|c|c|c|c|c|c|c|c|} 
 \hline
 & \multicolumn{6}{c|}{Base classes} & \multicolumn{6}{c|}{Novel classes} &\\
 \hline
Method & bathtub & bed & bookshelf & chair & desk & Average & dresser & nightstand & sofa & table & toilet & Average & All\\
 \hline
 SDCoT & 41.82 & 43.91 & 2.66 & \textbf{49.84} & 8.82 & 29.41 & \textbf{33.00} & 56.84 & 62.75 & 48.73 & 87.36 & 57.74 & 43.57\\ 
 \hline
 Ours & \textbf{52.62} & \textbf{62.31} & \textbf{6.01} & 48.47 & \textbf{12.66}& \textbf{36.41} & 26.67 & \textbf{61.82} & \textbf{63.82} & \textbf{51.63} & \textbf{90.04} & \textbf{58.80} & \textbf{47.60}\\ 
 \hline
\end{tabular}
}
\label{tab:per_class}
\end{table}

\begin{table}[t]
    \begin{minipage}{.5\linewidth}
      \centering
        \caption{Object detection performance mAP@0.25 of different augmentation techniques.}
        \label{tab:other_methods}
        \scalebox{0.75}{
        \begin{tabular}{|c|c|c|c|}
        \hline
        Method & Base & Novel & All\\\hline
        Mix3D~\cite{nekrasov2021mix3d} & 33.48 & 57.89 & 45.69 \\ \hline
        CutMix~\cite{yun2019cutmix} & 31.71 & 57.15 & 44.43 \\ \hline
        Ours & \textbf{36.41} & \textbf{58.80} & \textbf{47.60} \\ \hline
        \end{tabular}
        }
    \end{minipage}%
    \begin{minipage}{.5\linewidth}
      \centering
        \caption{Object detection performance mAP@0.25 with exclusion of components in our method.}
        \label{tab:ablation}
        \scalebox{0.75}{
        \begin{tabular}{|c|c|c|c|}
        \hline
        Method & Base & Novel & All\\\hline
        No cross-domain CP & 33.67 & \textbf{59.95} & 46.81 \\ 
         \hline
        No in-domain CP & 35.70 & 58.39 & 47.05 \\ 
         \hline
        No BN consistency & \textbf{36.52} & 57.20 & 46.86 \\ 
          \hline
        Ours & 36.41 & 58.80 & \textbf{47.60} \\ \hline
        \end{tabular}
        }
    \end{minipage} 
\end{table}

         

\subsection{Ablation Study}
To evaluate the effectiveness of the proposed augmentation method, we implemented two popular augmentation techniques for comparison,~\emph{i.e., } Mix3D~\cite{nekrasov2021mix3d} and CutMix~\cite{yun2019cutmix}. Mix3D combines $2$ point clouds into a mixed point cloud via concatenation, which results in excessive overlaps in the mixed point clouds. CutMix randomly replaces patches of point clouds with patches from other point clouds, which unnecessarily introduces extra context from the source domain. As reported in Table~\ref{tab:other_methods}, our method outperforms the two augmentation techniques. 
To evaluate the influence of the number of augmented data, we reported the model performance with different maximum numbers of augmented objects for cross-domain copy-paste,~\emph{i.e., } the mAP@0.25 scores are 47.60\% (1-2 objects), 47.08\% (1-3 objects), and 46.86\% (1-4
object). We can see that our model is not sensitive to changes. 

To evaluate the effectiveness of different components in our method for CIL under domain shift, we perform ablation studies by removing components in our approach. The results are demonstrated in Table~\ref{tab:ablation}. Without cross-domain copy-paste, the model achieves a lower result in base classes, which proves the effectiveness of cross-domain copy-paste augmentation in base class recognition. Similarly, it can be inferred from results without in-domain augmentation that in-domain copy-paste can enhance the model performance in novel classes. Moreover, statistics-level consistency provides around $0.8\%$ performance increment in all classes. Besides, by adding in-domain CP in SDCoT, the mAP@0.25 is increased to 46.60\%, and the score is increased to 46.86\% by adding both in-domain CP and cross-domain CP, which demonstrate the effectiveness of augmentation methods against domain shift. In summary, the results indicate the effectiveness of each component in our architecture.


\section{Conclusion}
In this work, we identify and explore a novel domain adaptive class-incremental learning paradigm for 3D object detection. To achieve both incremental learning and domain adaptation, we propose a novel 3D object detection framework, DA-CIL, in which we design a novel dual-domain copy-paste augmentation method to address both in-domain data scarcity and cross-domain distribution shift. We further enhance the dual-teacher knowledge distillation with multi-level consistency between different domains. Extensive experimental results and analysis demonstrate the superior performance of our method over baselines under the challenging class-incremental scenario. In our future work, we will explore more scenarios with different domain shifts, such as geography-to-geography, day-tonight, and simulation-to-reality to further verify the effectiveness of the proposed method.

\section*{Acknowledgment}
This research is supported by the Agency for Science, Technology and Research (A*STAR) under its AI3 HTPO Seed Fund  no. C211118008.

\clearpage

\bibliography{egbib}

\begin{thebibliography}{37}
\providecommand{\natexlab}[1]{#1}
\providecommand{\url}[1]{\texttt{#1}}
\expandafter\ifx\csname urlstyle\endcsname\relax
  \providecommand{\doi}[1]{doi: #1}\else
  \providecommand{\doi}{doi: \begingroup \urlstyle{rm}\Url}\fi

\bibitem[Cai et~al.(2021)Cai, Ravichandran, Maji, Fowlkes, Tu, and
  Soatto]{cai2021exponential}
Zhaowei Cai, Avinash Ravichandran, Subhransu Maji, Charless Fowlkes, Zhuowen
  Tu, and Stefano Soatto.
\newblock Exponential moving average normalization for self-supervised and
  semi-supervised learning.
\newblock In \emph{Proceedings of the IEEE/CVF Conference on Computer Vision
  and Pattern Recognition}, pages 194--203, 2021.

\bibitem[Chen and Chao(2021)]{chen2021gradual}
Hong-You Chen and Wei-Lun Chao.
\newblock Gradual domain adaptation without indexed intermediate domains.
\newblock \emph{Advances in Neural Information Processing Systems}, 2021.

\bibitem[Dai et~al.(2017)Dai, Chang, Savva, Halber, Funkhouser, and
  Nie{\ss}ner]{dai2017scannet}
Angela Dai, Angel~X Chang, Manolis Savva, Maciej Halber, Thomas Funkhouser, and
  Matthias Nie{\ss}ner.
\newblock Scannet: Richly-annotated 3d reconstructions of indoor scenes.
\newblock In \emph{Proceedings of the IEEE conference on computer vision and
  pattern recognition}, pages 5828--5839, 2017.

\bibitem[Dai et~al.(2021)Dai, Liu, Sun, Tong, Zhang, and Duan]{dai2021idm}
Yongxing Dai, Jun Liu, Yifan Sun, Zekun Tong, Chi Zhang, and Ling-Yu Duan.
\newblock Idm: An intermediate domain module for domain adaptive person re-id.
\newblock In \emph{Proceedings of the IEEE/CVF International Conference on
  Computer Vision}, 2021.

\bibitem[De~Lange et~al.(2021)De~Lange, Aljundi, Masana, Parisot, Jia,
  Leonardis, Slabaugh, and Tuytelaars]{de2021continual}
Matthias De~Lange, Rahaf Aljundi, Marc Masana, Sarah Parisot, Xu~Jia,
  Ale{\v{s}} Leonardis, Gregory Slabaugh, and Tinne Tuytelaars.
\newblock A continual learning survey: Defying forgetting in classification
  tasks.
\newblock \emph{IEEE transactions on pattern analysis and machine
  intelligence}, 44\penalty0 (7):\penalty0 3366--3385, 2021.

\bibitem[Dong et~al.(2021)Dong, Cong, Sun, Ma, and Wang]{dong2021i3dol}
Jiahua Dong, Yang Cong, Gan Sun, Bingtao Ma, and Lichen Wang.
\newblock I3dol: Incremental 3d object learning without catastrophic
  forgetting.
\newblock In \emph{Proceedings of the AAAI Conference on Artificial
  Intelligence}, pages 6066--6074, 2021.

\bibitem[Ghiasi et~al.(2021)Ghiasi, Cui, Srinivas, Qian, Lin, Cubuk, Le, and
  Zoph]{ghiasi2021simple}
Golnaz Ghiasi, Yin Cui, Aravind Srinivas, Rui Qian, Tsung-Yi Lin, Ekin~D Cubuk,
  Quoc~V Le, and Barret Zoph.
\newblock Simple copy-paste is a strong data augmentation method for instance
  segmentation.
\newblock In \emph{Proceedings of the IEEE/CVF Conference on Computer Vision
  and Pattern Recognition}, pages 2918--2928, 2021.

\bibitem[Hinton et~al.(2015)Hinton, Vinyals, Dean,
  et~al.]{hinton2015distilling}
Geoffrey Hinton, Oriol Vinyals, Jeff Dean, et~al.
\newblock Distilling the knowledge in a neural network.
\newblock \emph{arXiv preprint arXiv:1503.02531}, 2\penalty0 (7), 2015.

\bibitem[Hoffman et~al.(2018)Hoffman, Tzeng, Park, Zhu, Isola, Saenko, Efros,
  and Darrell]{hoffman2018cycada}
Judy Hoffman, Eric Tzeng, Taesung Park, Jun-Yan Zhu, Phillip Isola, Kate
  Saenko, Alexei Efros, and Trevor Darrell.
\newblock Cycada: Cycle-consistent adversarial domain adaptation.
\newblock In \emph{International conference on machine learning}, pages
  1989--1998. Pmlr, 2018.

\bibitem[Hou et~al.(2019)Hou, Pan, Loy, Wang, and Lin]{hou2019learning}
Saihui Hou, Xinyu Pan, Chen~Change Loy, Zilei Wang, and Dahua Lin.
\newblock Learning a unified classifier incrementally via rebalancing.
\newblock In \emph{Proceedings of the IEEE/CVF Conference on Computer Vision
  and Pattern Recognition}, pages 831--839, 2019.

\bibitem[Jie et~al.(2022)Jie, Liang, Zhao, Chen, Chang, and Zeng]{jie2022adan}
Luyang Jie, Pengchen Liang, Ziyuan Zhao, Jianguo Chen, Qing Chang, and Zeng
  Zeng.
\newblock Adan: An adversarial domain adaptation neural network for early
  gastric cancer prediction.
\newblock In \emph{2022 44th Annual International Conference of the IEEE
  Engineering in Medicine \& Biology Society (EMBC)}, pages 2169--2172. IEEE,
  2022.

\bibitem[Kirkpatrick et~al.(2017)Kirkpatrick, Pascanu, Rabinowitz, Veness,
  Desjardins, Rusu, Milan, Quan, Ramalho, Grabska-Barwinska,
  et~al.]{kirkpatrick2017overcoming}
James Kirkpatrick, Razvan Pascanu, Neil Rabinowitz, Joel Veness, Guillaume
  Desjardins, Andrei~A Rusu, Kieran Milan, John Quan, Tiago Ramalho, Agnieszka
  Grabska-Barwinska, et~al.
\newblock Overcoming catastrophic forgetting in neural networks.
\newblock \emph{Proceedings of the national academy of sciences}, 114\penalty0
  (13):\penalty0 3521--3526, 2017.

\bibitem[Lopez-Paz and Ranzato(2017)]{lopez2017gradient}
David Lopez-Paz and Marc'Aurelio Ranzato.
\newblock Gradient episodic memory for continual learning.
\newblock \emph{Advances in neural information processing systems}, 30, 2017.

\bibitem[Luo et~al.(2021)Luo, Cai, Zhou, Zhang, Zhao, Yi, Lu, Li, Zhang, and
  Liu]{luo2021unsupervised}
Zhipeng Luo, Zhongang Cai, Changqing Zhou, Gongjie Zhang, Haiyu Zhao, Shuai Yi,
  Shijian Lu, Hongsheng Li, Shanghang Zhang, and Ziwei Liu.
\newblock Unsupervised domain adaptive 3d detection with multi-level
  consistency.
\newblock In \emph{Proceedings of the IEEE/CVF International Conference on
  Computer Vision}, pages 8866--8875, 2021.

\bibitem[Mallya and Lazebnik(2018)]{mallya2018packnet}
Arun Mallya and Svetlana Lazebnik.
\newblock Packnet: Adding multiple tasks to a single network by iterative
  pruning.
\newblock In \emph{Proceedings of the IEEE conference on Computer Vision and
  Pattern Recognition}, pages 7765--7773, 2018.

\bibitem[Nekrasov et~al.(2021)Nekrasov, Schult, Litany, Leibe, and
  Engelmann]{nekrasov2021mix3d}
Alexey Nekrasov, Jonas Schult, Or~Litany, Bastian Leibe, and Francis Engelmann.
\newblock Mix3d: Out-of-context data augmentation for 3d scenes.
\newblock In \emph{2021 International Conference on 3D Vision (3DV)}, pages
  116--125. IEEE, 2021.

\bibitem[Nwe et~al.(2020)Nwe, Min, Gopalakrishnan, Lin, Prasad, Dong, Li, and
  Pahwa]{nwe2020improving}
Tin~Lay Nwe, Oo~Zaw Min, Saisubramaniam Gopalakrishnan, Dongyun Lin, Shitala
  Prasad, Sheng Dong, Yiqun Li, and Ramanpreet~Singh Pahwa.
\newblock Improving 3d brain tumor segmentation with predict-refine mechanism
  using saliency and feature maps.
\newblock In \emph{2020 IEEE International Conference on Image Processing
  (ICIP)}, pages 2671--2675. IEEE, 2020.

\bibitem[Pahwa et~al.(2021)Pahwa, Nwe, Chang, Min, Jie, Gopalakrishnan, Wee,
  Qin, Rao, Dai, et~al.]{pahwa2021automated}
Ramanpreet~Singh Pahwa, Ma~Tin~Lay Nwe, Richard Chang, Oo~Zaw Min, Wang Jie,
  Saisubramaniam Gopalakrishnan, David Ho~Soon Wee, Ren Qin, Vempati~Srinivasa
  Rao, Haiwen Dai, et~al.
\newblock Automated attribute measurements of buried package features in 3d
  x-ray images using deep learning.
\newblock In \emph{2021 IEEE 71st Electronic Components and Technology
  Conference (ECTC)}, pages 2196--2204. IEEE, 2021.

\bibitem[Pham et~al.(2020)Pham, Sevestre, Pahwa, Zhan, Pang, Chen, Mustafa,
  Chandrasekhar, and Lin]{pham20203d}
Quang-Hieu Pham, Pierre Sevestre, Ramanpreet~Singh Pahwa, Huijing Zhan, Chun~Ho
  Pang, Yuda Chen, Armin Mustafa, Vijay Chandrasekhar, and Jie Lin.
\newblock A* 3d dataset: Towards autonomous driving in challenging
  environments.
\newblock In \emph{2020 IEEE International Conference on Robotics and
  Automation (ICRA)}, pages 2267--2273. IEEE, 2020.

\bibitem[Qi et~al.(2017{\natexlab{a}})Qi, Su, Mo, and Guibas]{qi2017pointnet}
Charles~R Qi, Hao Su, Kaichun Mo, and Leonidas~J Guibas.
\newblock Pointnet: Deep learning on point sets for 3d classification and
  segmentation.
\newblock In \emph{Proceedings of the IEEE conference on computer vision and
  pattern recognition}, pages 652--660, 2017{\natexlab{a}}.

\bibitem[Qi et~al.(2019)Qi, Litany, He, and Guibas]{qi2019deep}
Charles~R Qi, Or~Litany, Kaiming He, and Leonidas~J Guibas.
\newblock Deep hough voting for 3d object detection in point clouds.
\newblock In \emph{proceedings of the IEEE/CVF International Conference on
  Computer Vision}, pages 9277--9286, 2019.

\bibitem[Qi et~al.(2017{\natexlab{b}})Qi, Yi, Su, and Guibas]{qi2017pointnet++}
Charles~Ruizhongtai Qi, Li~Yi, Hao Su, and Leonidas~J Guibas.
\newblock Pointnet++: Deep hierarchical feature learning on point sets in a
  metric space.
\newblock \emph{Advances in neural information processing systems}, 30,
  2017{\natexlab{b}}.

\bibitem[Ramamonjison et~al.(2021)Ramamonjison, Banitalebi-Dehkordi, Kang, Bai,
  and Zhang]{ramamonjison2021simrod}
Rindra Ramamonjison, Amin Banitalebi-Dehkordi, Xinyu Kang, Xiaolong Bai, and
  Yong Zhang.
\newblock Simrod: A simple adaptation method for robust object detection.
\newblock In \emph{Proceedings of the IEEE/CVF International Conference on
  Computer Vision}, pages 3570--3579, 2021.

\bibitem[Rebuffi et~al.(2017)Rebuffi, Kolesnikov, Sperl, and
  Lampert]{rebuffi2017icarl}
Sylvestre-Alvise Rebuffi, Alexander Kolesnikov, Georg Sperl, and Christoph~H
  Lampert.
\newblock icarl: Incremental classifier and representation learning.
\newblock In \emph{Proceedings of the IEEE conference on Computer Vision and
  Pattern Recognition}, pages 2001--2010, 2017.

\bibitem[Serra et~al.(2018)Serra, Suris, Miron, and
  Karatzoglou]{serra2018overcoming}
Joan Serra, Didac Suris, Marius Miron, and Alexandros Karatzoglou.
\newblock Overcoming catastrophic forgetting with hard attention to the task.
\newblock In \emph{International Conference on Machine Learning}, pages
  4548--4557. PMLR, 2018.

\bibitem[Song et~al.(2015)Song, Lichtenberg, and Xiao]{song2015sun}
Shuran Song, Samuel~P Lichtenberg, and Jianxiong Xiao.
\newblock Sun rgb-d: A rgb-d scene understanding benchmark suite.
\newblock In \emph{Proceedings of the IEEE conference on computer vision and
  pattern recognition}, pages 567--576, 2015.

\bibitem[Tarvainen and Valpola(2017)]{tarvainen2017mean}
Antti Tarvainen and Harri Valpola.
\newblock Mean teachers are better role models: Weight-averaged consistency
  targets improve semi-supervised deep learning results.
\newblock \emph{Advances in neural information processing systems}, 30, 2017.

\bibitem[Tsai et~al.(2018)Tsai, Hung, Schulter, Sohn, Yang, and
  Chandraker]{tsai2018learning}
Yi-Hsuan Tsai, Wei-Chih Hung, Samuel Schulter, Kihyuk Sohn, Ming-Hsuan Yang,
  and Manmohan Chandraker.
\newblock Learning to adapt structured output space for semantic segmentation.
\newblock In \emph{Proceedings of the IEEE conference on computer vision and
  pattern recognition}, pages 7472--7481, 2018.

\bibitem[Wang and Deng(2018)]{wang2018deep}
Mei Wang and Weihong Deng.
\newblock Deep visual domain adaptation: A survey.
\newblock \emph{Neurocomputing}, 312:\penalty0 135--153, 2018.

\bibitem[Wang et~al.(2020)Wang, Chen, You, Li, Hariharan, Campbell, Weinberger,
  and Chao]{wang2020train}
Yan Wang, Xiangyu Chen, Yurong You, Li~Erran Li, Bharath Hariharan, Mark
  Campbell, Kilian~Q Weinberger, and Wei-Lun Chao.
\newblock Train in germany, test in the usa: Making 3d object detectors
  generalize.
\newblock In \emph{Proceedings of the IEEE/CVF Conference on Computer Vision
  and Pattern Recognition}, pages 11713--11723, 2020.

\bibitem[Yang et~al.(2021)Yang, Shi, Wang, Li, and Qi]{yang2021st3d}
Jihan Yang, Shaoshuai Shi, Zhe Wang, Hongsheng Li, and Xiaojuan Qi.
\newblock St3d: Self-training for unsupervised domain adaptation on 3d object
  detection.
\newblock In \emph{Proceedings of the IEEE/CVF Conference on Computer Vision
  and Pattern Recognition}, pages 10368--10378, 2021.

\bibitem[Yun et~al.(2019)Yun, Han, Oh, Chun, Choe, and Yoo]{yun2019cutmix}
Sangdoo Yun, Dongyoon Han, Seong~Joon Oh, Sanghyuk Chun, Junsuk Choe, and
  Youngjoon Yoo.
\newblock Cutmix: Regularization strategy to train strong classifiers with
  localizable features.
\newblock In \emph{Proceedings of the IEEE/CVF international conference on
  computer vision}, pages 6023--6032, 2019.

\bibitem[Zhao and Lee(2022)]{zhao2022static}
Na~Zhao and Gim~Hee Lee.
\newblock Static-dynamic co-teaching for class-incremental 3d object detection.
\newblock In \emph{Proceedings of the AAAI Conference on Artificial
  Intelligence}, pages 3436--3445, 2022.

\bibitem[Zhao et~al.(2020)Zhao, Chua, and Lee]{zhao2020sess}
Na~Zhao, Tat-Seng Chua, and Gim~Hee Lee.
\newblock Sess: Self-ensembling semi-supervised 3d object detection.
\newblock In \emph{Proceedings of the IEEE/CVF Conference on Computer Vision
  and Pattern Recognition}, pages 11079--11087, 2020.

\bibitem[Zhao et~al.(2021)Zhao, Xu, Li, Zeng, and Guan]{zhao2021mt}
Ziyuan Zhao, Kaixin Xu, Shumeng Li, Zeng Zeng, and Cuntai Guan.
\newblock Mt-uda: Towards unsupervised cross-modality medical image
  segmentation with limited source labels.
\newblock In \emph{International Conference on Medical Image Computing and
  Computer-Assisted Intervention}, pages 293--303. Springer, 2021.

\bibitem[Zhao et~al.(2022)Zhao, Zhou, Zeng, Guan, and Zhou]{zhao2022meta}
Ziyuan Zhao, Fangcheng Zhou, Zeng Zeng, Cuntai Guan, and S~Kevin Zhou.
\newblock Meta-hallucinator: Towards few-shot cross-modality cardiac image
  segmentation.
\newblock In \emph{International Conference on Medical Image Computing and
  Computer-Assisted Intervention}, pages 128--139. Springer, 2022.

\bibitem[Zou et~al.(2019)Zou, Yu, Liu, Kumar, and Wang]{zou2019confidence}
Yang Zou, Zhiding Yu, Xiaofeng Liu, BVK Kumar, and Jinsong Wang.
\newblock Confidence regularized self-training.
\newblock In \emph{Proceedings of the IEEE/CVF International Conference on
  Computer Vision}, pages 5982--5991, 2019.

\end{thebibliography}
\end{document}


\maketitle

\maketitle

\begin{figure*}[htb]
\centering
\includegraphics[width=0.7\textwidth]{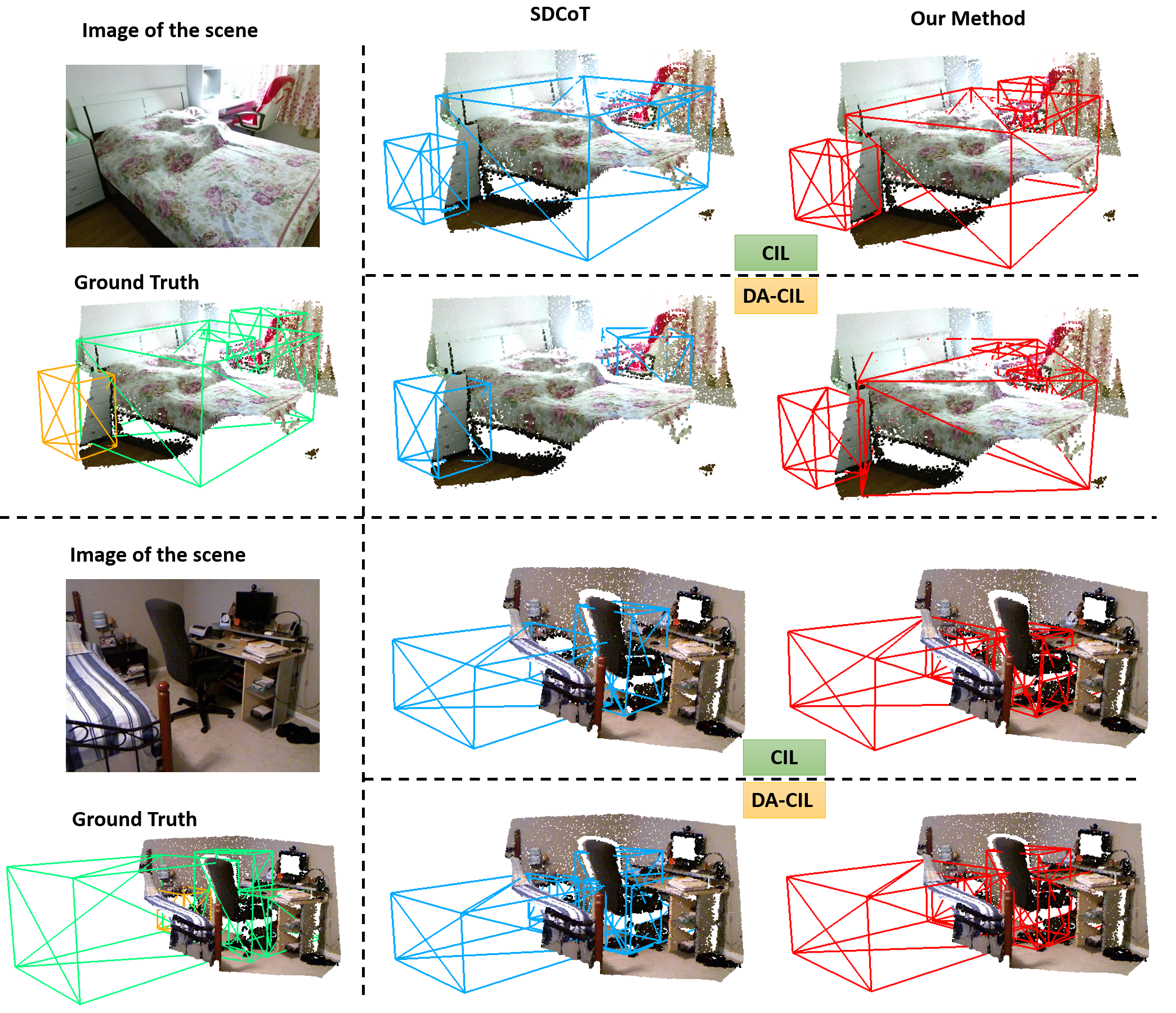}
\caption{Qualitative results on SUN RGB-D validation set. \textcolor{green}{Green} and \textcolor{yellow}{Yellow} represent ground-truth annotations of base and novel classes, respectively. CIL refers to models trained on the same domain and DA-CIL refers to models trained on source and target domains with domain shift.}
\label{fig:visualization}
\end{figure*}

In this supplementary material, we provide visualizations of inference results with our proposed approach and compare it with the baseline method. As illustrated in Fig.~\ref{fig:visualization}, our method is able to accurately detect both old and novel classes in the target domain, overcoming the domain shift in old classes. Moreover, our model consistently outperforms the baseline SDCoT in both scenarios,~\emph{i.e.}, class-incremental learning (CIL), and domain adaptive class-incremental learning (DA-CIL). Besides, our method identifies partially occluded and cluttered objects, which are very challenging targets.

\begin{figure*}[htb]
\centering
\includegraphics[width=0.7\textwidth]{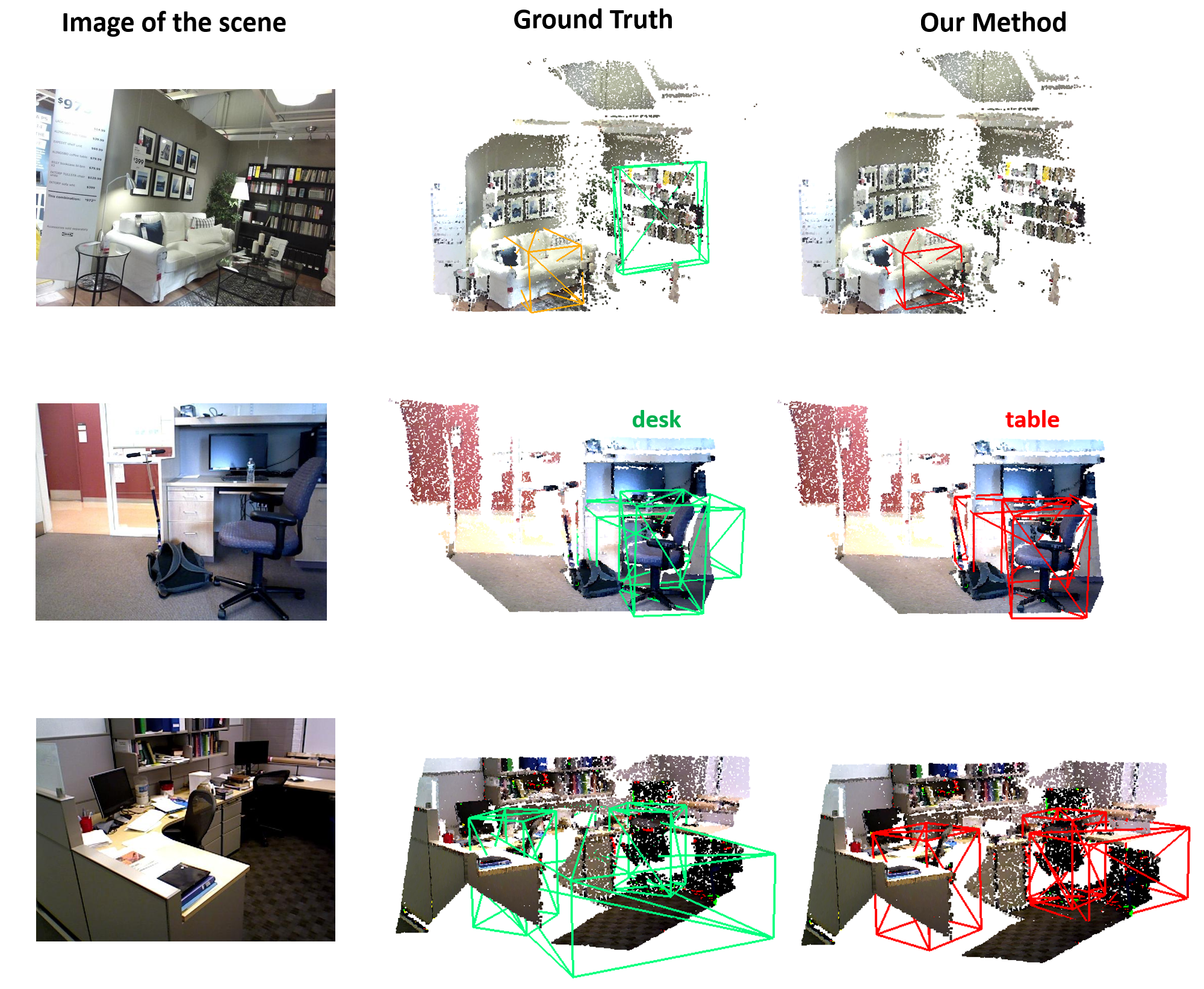}
\caption{Failure cases on SUN RGB-D validation set under the DA-CIL setting. \textcolor{green}{Green} and \textcolor{yellow}{Yellow} represent ground-truth annotations of base and novel classes, respectively. We show three examples from top to bottom.}
\label{fig:failure}
\end{figure*}

In Fig.~\ref{fig:failure}, we present failure examples of our approach on the SUN RGB-D validation set. In the first row of Fig.~\ref{fig:failure}, our method fails to detect the bookshelf in the point clouds. The failure is due to the loss of geometric structure of the bookshelf in the point cloud data, which is partially occluded by books and other objects. Besides, the gap between source and target domains in the bookshelf class is also an obstacle to accurate detection. 
In the second row of Fig.~\ref{fig:failure}, the desk object is misclassified as a table, which is likely due to geometric similarities between the $2$ classes. In the last row of Fig.~\ref{fig:failure}, the bounding box of the desk object deviates from the ground-truth, which can be attributed to the large size and irregular shape of the object.

